\pdfoutput=1

\documentclass[11pt]{article}

\usepackage{emnlp2021}

\usepackage{times}
\usepackage{latexsym}
\usepackage{array}
\newcolumntype{P}[1]{>{\centering\arraybackslash}p{#1}}
\newcolumntype{M}[1]{>{\centering\arraybackslash}m{#1}}

\usepackage{graphicx}
\usepackage{svg}
\usepackage{comment}
\usepackage{booktabs}
\usepackage{array}
\usepackage{caption}
\usepackage{subcaption}
\usepackage{tablefootnote}
\newcommand{\PreserveBackslash}[1]{\let\temp=\\#1\let\\=\temp}
\newcolumntype{C}[1]{>{\PreserveBackslash\centering}p{#1}}
\newcolumntype{R}[1]{>{\PreserveBackslash\raggedleft}p{#1}}
\newcolumntype{L}[1]{>{\PreserveBackslash\raggedright}p{#1}}

\usepackage{cellspace}
\usepackage{array}
\newcolumntype{H}{>{\setbox0=\hbox\bgroup}c<{\egroup}@{}}

\makeatletter
\def\thickhline{%
  \noalign{\ifnum0=`}\fi\hrule \@height \thickarrayrulewidth \futurelet
   \reserved@a\@xthickhline}
\def\@xthickhline{\ifx\reserved@a\thickhline
               \vskip\doublerulesep
               \vskip-\thickarrayrulewidth
             \fi
      \ifnum0=`{\fi}}
\makeatother

\newlength{\thickarrayrulewidth}
\usepackage{microtype}
\usepackage{multirow}
\usepackage{multicol}

\newcommand\pubmed{{\textsc{PubMedM\&M}}}
\newcommand\chemsyn{{\textsc{ChemSyn}}}
\newcommand\wlp{{\textsc{WLP}}}
\newcommand\xwlp{{\textsc{XWLP}}}
\newcommand\chemu{{\textsc{ChEMU}}}
\newcommand\recipe{{\textsc{Recipe}}}

\newcommand\roberta{{RoBERTa}}
\newcommand\procbert{{ProcBERT}}
\newcommand\procroberta{{Proc-RoBERTa}}
\newcommand\scibert{{SciBERT}}

\newcommand\bertbase{{BERT$_{base}$}}
\newcommand\bertlarge{{BERT$_{large}$}}
\newcommand\robbase{{RoBERTa$_{base}$}}
\newcommand\roblarge{{RoBERTa$_{large}$}}

\newcommand\biomedrob{{BioMed-RoBERTa}}

\newcommand\easyadapt{{EasyAdapt}}

\newcommand{\quotes}[1]{``#1''}

\usepackage[T1]{fontenc}

\usepackage[utf8]{inputenc}

\usepackage{microtype}

\usepackage{scalerel,xparse}
\NewDocumentCommand\emojibert{}{
    \includegraphics[scale=0.18]{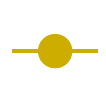}
}
\NewDocumentCommand\emojiprocbert{}{
    \includegraphics[scale=0.18]{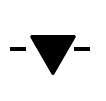}
}

\title{Pre-train or Annotate? Domain Adaptation with a Constrained Budget}

\author{Fan Bai, Alan Ritter, Wei Xu \\
  School of Interactive Computing \\
  Georgia Institute of Technology \\
 \texttt{\{fan.bai, alan.ritter, wei.xu\}@cc.gatech.edu} 
\\}

\begin{document}
\maketitle
\begin{abstract}

Recent work has demonstrated that pre-training in-domain language models can boost performance when adapting to a new domain. However, the costs associated with pre-training raise an important question: given a fixed budget, what steps should an NLP practitioner take to maximize performance? In this paper, we view domain adaptation with a constrained budget as a consumer choice problem, where the goal is to select an optimal combination of data annotation and pre-training.  We measure annotation costs of three procedural text datasets, along with the pre-training costs of several in-domain language models.  The utility of different combinations of pre-training and data annotation are evaluated under varying budget constraints to assess which combination strategy works best. We find that for small budgets, spending all funds on annotation leads to the best performance; once the budget becomes large enough, however, a combination of data annotation and in-domain pre-training yields better performance. Our experiments suggest task-specific data annotation should be part of an economical strategy when adapting an NLP model to a new domain.\footnote{Our code and data are publicly available on Github: \url{https://github.com/bflashcp3f/ProcBERT}.}

\end{abstract}

\section{Introduction}
\label{sec:intro}
The conventional wisdom on semi-supervised learning and unsupervised domain adaptation is that labeled data is expensive; therefore, training on a combination of labeled and unlabeled data is an economical approach to improve performance when adapting to a new domain \citep{blum1998combining,daume2006domain,hoffman2018cycada,chen2020mixtext}.
\begin{figure}[ht]
\includegraphics[width=0.96\columnwidth]{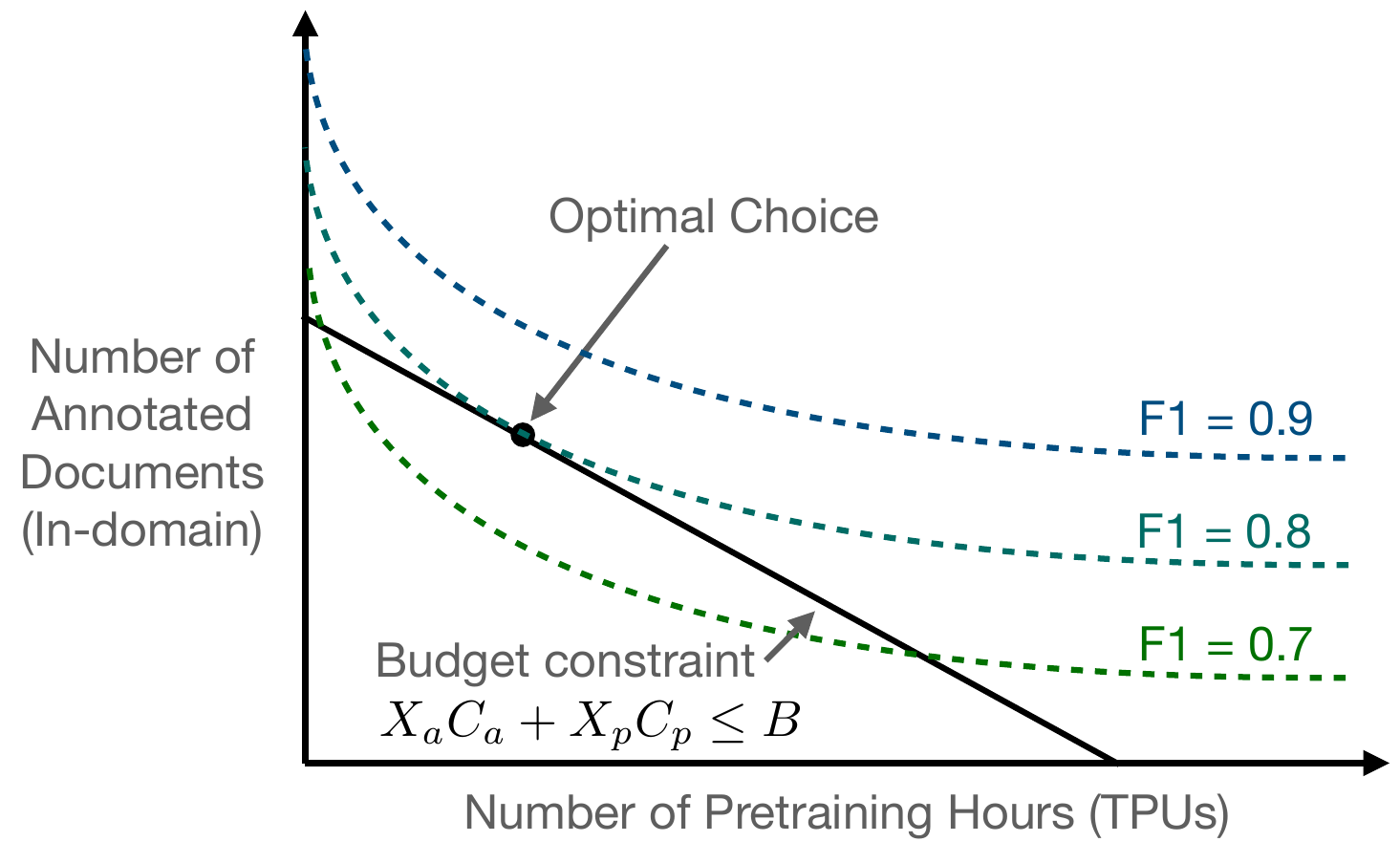}
\caption{\label{fig:consumer_choice} We view domain adaptation as a consumer choice problem \citep{becker1965theory,lancaster1966new}. The NLP practitioner (consumer) is faced with the problem of choosing an optimal combination of annotation and pre-training under a constrained budget.  This figure is purely for illustration and is not based on experimental data. 
}
\vspace{-.2cm}
\end{figure}
Recent work has shown that pre-training in-domain Transformers is an effective method for unsupervised adaptation \cite{Han2019UnsupervisedDA,wright2020transformer} 
and even boosts performance when large quantities of in-domain data are available \cite{Gururangan2020DontSP}.  
However, modern pre-training methods incur substantial costs \citep{izsak2021train}, and generate carbon emissions \citep{Strubell2019EnergyAP,schwartz2019green,bender2021dangers}. 
This raises an important question: given a fixed budget to improve a model's performance, what steps should an NLP practitioner take?  On one hand, they could hire annotators to label in-domain task-specific data, while on the other, they could buy or rent GPUs or TPUs to pre-train large in-domain language models.  In this paper, we empirically study the best strategy for adapting to a new domain given a fixed budget.

We view the NLP practitioner’s dilemma of how to adapt to a new domain as a problem of consumer choice, a classical problem in microeconomics \citep{becker1965theory,lancaster1966new}.  
As illustrated in Figure \ref{fig:consumer_choice}, the NLP practitioner (consumer) can obtain $X_a$ annotated documents (by hiring annotators) at a cost of $C_a$ each, and $X_p$ hours of pre-training (by renting GPUs or TPUs) at a cost of $C_p$ per hour.  Given a fixed budget $B$, the consumer may choose any combination that fits within the budget constraint $X_a C_a + X_p C_P \leq B$.  The goal is to choose a combination that maximizes the utility function, $U(X_a, X_p)$, which can be defined using an appropriate performance metric, such as $F_1$ score, that is achieved after pre-training for $X_p$ hours and then fine-tuning on $X_a$ in-domain documents. 


To empirically estimate the cost of annotation, we hire annotators to label domain-specific documents for supervised fine-tuning in three procedural text domains: wet-lab protocols, paragraphs describing scientific procedures in PubMed articles, and chemical synthesis procedures described in patents.  We choose to target natural language understanding for scientific procedures in this study, because there is an opportunity to help automate lab protocols and support more reproducible scientific experiments, yet few annotated datasets currently exist in these domains.  Furthermore, annotation of scientific procedures is not easily amenable to crowdsourcing, making this an ideal testbed for pre-training-based domain adaptation. We measure the cost of in-domain pre-training on a large collection of unlabeled procedural texts using Google's Cloud TPUs.\footnote{\url{https://cloud.google.com/tpu}}
Model performance is then evaluated under varying budget constraints in six source and target domain combinations.



Our analysis suggests that given current costs of pre-training large Transformer models, such as BERT \citep{devlin-etal-2019-bert}, and RoBERTa \citep{Liu2019RoBERTaAR}, in-domain data annotation should always be part of an economical strategy when adapting a single NLP system to a new domain.
For small budgets (e.g. less than \$800 USD), spending all funds on annotation is the best policy; however, as more funding becomes available, a combination of pre-training and annotation is the best choice.    

This paper addresses a specific question that is often faced by NLP practitioners working on applications: what is the most economical approach to adapt an NLP system to a new domain when no pre-trained models or task-annotated datasets are initially available?  If multiple NLP systems need to be adapted to a single target domain, model costs can be amortized, making pre-training an attractive option for smaller budgets.

\section{Scope of the Study}
\label{sec:assumptions}
In this study, we focus on a typical scenario faced by an NLP practitioner to adapt a single NLP system to a single new domain, maximizing performance within a constrained budget.
We consider only the direct benefit on the target task in our main analysis (\S \ref{sec:measure_utility}), however we do provide additional analysis of positive externalities on other related tasks that may benefit from a new pre-trained model in \S \ref{sec:other_procedural}.

We estimate cost based on two major expenses: annotating task-specific data (\S \ref{sec:estimate_c_a}) and pre-training domain-specific models using TPUs (\S \ref{sec:estimate_c_p}). 
Note that fine-tuning costs are not included in our analysis, as they are nearly equal whether budget is invested into pre-training or annotation.\footnote{These are also not a significant portion of overall costs (we estimate \$1.95 based on Google Cloud rates for P100 GPUs.)}
We assume a generic BERT is the closest zero-cost model that is initially available, which is likely the case in real-world domain adaptation scenarios (especially for non-English languages). Our experiments are designed to simulate a scenario where no domain-specific model is initially available. 
We also assume that the NLP engineer's salary is a fixed cost; in other words, their salary will be the same whether they spend time pre-training models or managing a group of annotators.\footnote{Based on our experience, pre-training in-domain models (collecting training corpus is non-trivial) and managing a team of annotators are roughly comparable in terms of effort.} Our primary concerns are about financial and environmental costs, rather than the overall time needed to obtain the adapted model. If the timeline is an important factor, the annotation process can be possibly sped up by hiring more annotators.

\section{Estimating Annotation Cost ($C_a$)}
\label{sec:estimate_c_a}

In this section, we present our estimates of the annotation cost for three procedural text datasets from specialized scientific domains, which enable a comparison of model performance under varying budget constraints (\S \ref{sec:utility_result}).

\begin{table*}[t!]
\small
\renewcommand{\arraystretch}{1.1}
\begin{center}
\setlength{\tabcolsep}{3pt}
\resizebox{0.96\textwidth}{!}{\begin{tabular}{p{3cm}lccrccccc}
\toprule \textbf{Dataset} & \textbf{Domain} & \textbf{Task} & \textbf{\#Files (train/dev/test)} & \multicolumn{1}{c}{\textbf{\#Sentences}} & \multicolumn{1}{c}{\textbf{\#Cases}} & \textbf{\#Classes} & \textbf{Total Cost} & \textbf{Price/File} & \textbf{Price/Sent.}\\ \midrule
    & \multirow{2}{*}{biology}
                                 & NER    & \multirow{2}{*}{726 (492/123/111)} & \multirow{2}{*}{17,658{ }{ }{ }{ }} & 185,313 & 20 &  \multirow{2}{*}{\$7,820} & \multirow{2}{*}{\$10.8} & \multirow{2}{*}{\$0.44}\\
 & & RE     &                      &                         & 124,803 & 16 & &  \\
 
\multirow{1}{*}{\begin{minipage}{.3\textwidth} \vspace{-1.1cm} 
\textsc{WLP} \newline
\cite{Tabassum2020WNUT2020T1}\end{minipage}}  & \multicolumn{9}{l}{ \quad\quad\quad\quad\quad\quad\quad\quad\quad\begin{minipage}{.3\textwidth}\vspace{-.1cm} \includegraphics[height=14mm]{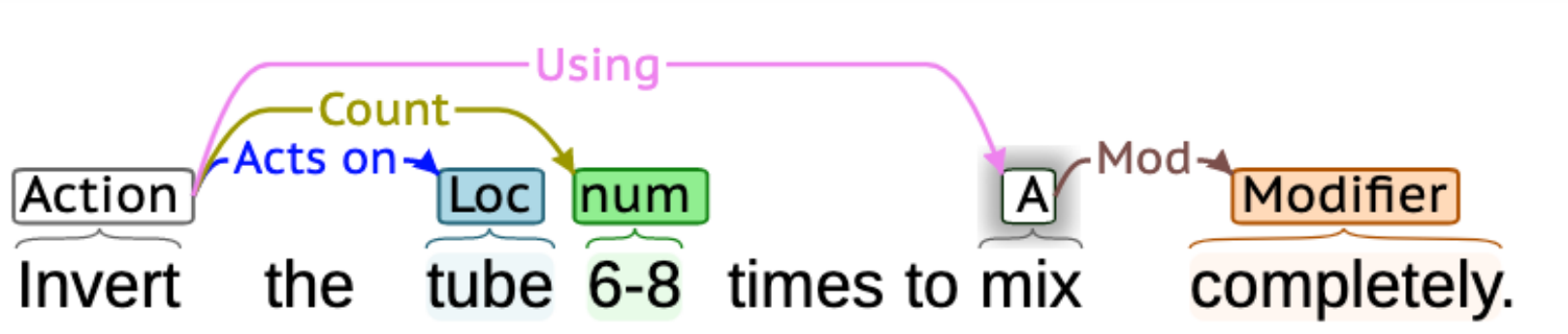} \end{minipage}} \\
 
\midrule

      & \multirow{2}{*}{biomed}
                                 & NER    & \multirow{2}{*}{191 (100/41/50)} & \multirow{2}{*}{1,699{ }{ }{ }{ }}  & 12,131 & 24 & \multirow{2}{*}{\$1,730} & \multirow{2}{*}{\$9.1} & \multirow{2}{*}{\$1.02} \\
\multirow{1}{*}{\begin{minipage}{.3\textwidth} \vspace{0.3cm} 
\pubmed{} \newline (this work)
\end{minipage}}                               & & RE     &                      &                         & 8,987   & 17 & &  \\ 

 \multicolumn{10}{l}{\quad\quad\quad\quad\quad\quad\quad \begin{minipage}{.3\textwidth}\vspace{-.1cm} \includegraphics[height=14mm]{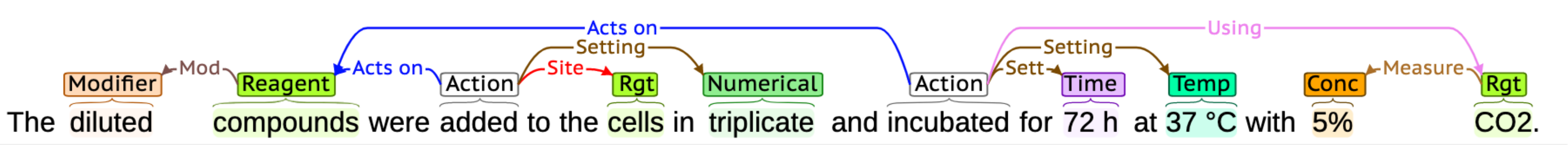} \end{minipage}} \\

\midrule

     & \multirow{2}{*}{chemistry}
                                 & NER    & \multirow{2}{*}{992 (793/99/100)} & \multirow{2}{*}{8,331{ }{ }{ }{ }}  & 53,423     & 24  & \multirow{2}{*}{\$5,000} & \multirow{2}{*}{\$4.7} & \multirow{2}{*}{\$0.60} \\
\multirow{1}{*}{\begin{minipage}{.3\textwidth} \vspace{0.3cm} 
\chemsyn{} \newline (this work)
\end{minipage}}                                 & & RE     &                      &                         & 46,878       & 17 & &  \\ 

 \multicolumn{10}{l}{\quad\quad\quad\quad\quad\quad\quad\quad\quad \begin{minipage}{.3\textwidth}\vspace{-.1cm} \includegraphics[height=14mm]{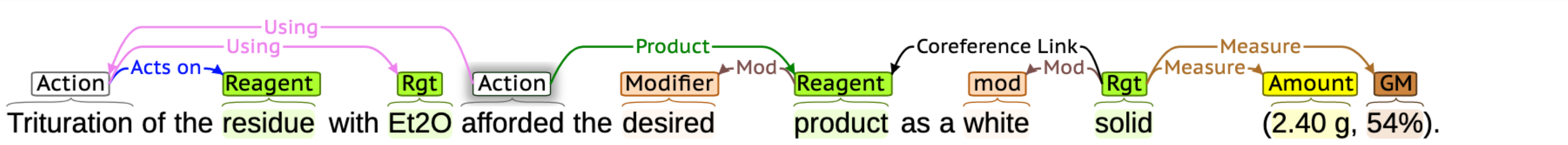} \end{minipage}} \\

\bottomrule
\end{tabular}}
\end{center}
\vspace{-.2cm}
\caption{\label{tab:task_statistics}Statistics and examples of three procedural text datasets.}
\end{table*}


\paragraph{Annotated Procedural Text Datasets.}
We experiment with three procedural text corpora, including Wet Lab Protocols \cite[WLP;][]{Tabassum2020WNUT2020T1} and two new datasets we created for this study, which include scientific articles and chemical patents. Statistics of the three datasets are shown in Table \ref{tab:task_statistics}. The \textbf{WLP corpus} includes 726 wet lab experiment instructions collected from \url{protocols.io}
which are annotated using an inventory of
20 entity types and 16 relation types. Following the same annotation scheme, we annotate \textsc{PubMedM\&M} and \textsc{ChemSyn}. The \textbf{\pubmed{} corpus} consists of 191 double-annotated experimental paragraphs extracted from the \textit{Materials and Methods} section of PubMed articles. 
The \textbf{\chemsyn{} corpus} consists of 992 chemical synthesis procedures described in patents, 500 of which are double-annotated. Unlike the succinct, informal language style in \wlp{}, \pubmed{} represents an academic writing style, as it comes from published research papers (see Table \ref{tab:task_statistics}). 
More details on data pre-processing, annotation and inter-annotator agreement scores can be found in Appendix \ref{sec:data_anno} 





\paragraph{Annotation Cost.}
We recruit undergraduate students to annotate the datasets using the BRAT annotation tool.\footnote{\url{https://github.com/nlplab/brat}}
Annotators are paid 13 USD / hour throughout the process, which is the standard rate for undergraduate students at our university.  Estimates of the cost of annotation, $C_a$, per-sentence are presented in Table \ref{tab:task_statistics}.\footnote{The average time to annotate each sentence varies across datasets based on the complexity of the text, average length of sentences, etc.}

\section{Estimating Pre-training Cost ($C_p$)}
\label{sec:estimate_c_p}

To evaluate varied strategies for combining pretraining and annotation given a fixed budget, we need accurate estimates on the cost of annotation, $C_a$, and pretraining, $C_p$.
Having estimated the cost of annotating in-domain procedural text corpora in \S \ref{sec:estimate_c_a}, we now turn to estimate the cost of in-domain pretraining.  
Specifically, we consider two popular approaches: 1) training an in-domain language model from scratch; 2) continued pre-training using an off-the-shelf model. 

\paragraph{\textsc{Procedure} Corpus Collection.}
To pre-train our models, we create a novel collection of procedural texts from the same domains as the annotated data in \S\ref{sec:estimate_c_a}, hereinafter referred to as the \textsc{Procedure} corpus.  

Specially trained classifiers were used to identify paragraphs describing experimental procedures. For PubMed, a classifier was used to identify paragraphs describing experimental procedures by fine-tuning \scibert{} \citep{Beltagy2019SciBERT} on the SciSeg dataset \citep{Dasigi2017ExperimentSI}, which is annotated with scientific discourse structure, to extract procedures from the {\em Materials and Methods} section of 680k articles. 
For the chemical synthesis domain, the chemical reaction extractor developed by \citet{Lowe2012ExtractionOC} was applied to the {\em Description} section of 303k patents (174k U.S. and 129k European) we collected from USPTO\footnote{\url{https://www.uspto.gov/}} and EPO\footnote{\url{https://www.epo.org/}}.
More details of our data collection process can be found in Appendix \ref{sec:app_corpus}. 

Cooking recipes are also an important domain for research on procedural text understanding, therefore we include the text component of the Recipe1M+ dataset \citep{Marn2019Recipe1MAD} in the \textsc{Procedure} pre-training corpus. In total, our \textbf{\textsc{Procedure}} collection contains around 1.1 billion words; more statistics are shown in Table \ref{tab:corpus_size}. 
In addition, we create an extended version, \textbf{\textsc{Procedure+}}, consisting of 12 billion words, where we up-sample the procedural paragraphs 6 times and combine them with the original full text of 680k PubMed articles and 303k chemical patents.  This up-sampling ensures at least half of the text is procedural.

\paragraph{Pre-training Process and Cost.}
\label{sec:pre-training_cost}

We train two procedural domain language models on the Google Cloud Platform using 8-core v3 TPUs:
1) \textbf{\procbert{}}, a \bertbase{} model pre-trained from scratch using our \textsc{Procedure+} corpus, and
2) \textbf{\procroberta{}}, for which we continued pre-training \robbase{} on the \textsc{Procedure} corpus following \citet{Gururangan2020DontSP}.

We pre-train \procbert{} using the TensorFlow codebase of BERT.\footnote{\url{https://github.com/google-research/bert}} Following \citet{devlin-etal-2019-bert}, we deploy a two-step regime: the model is trained with sequence length 128 and batch size 512 for 1 million steps at a rate of 4.71 steps/second. Then, it is trained for 100k more steps using sequences of length 512 and a batch size of 256 at a rate of 1.83 steps/second. The pretraining process takes about
74 hours, and the total cost is about 620 USD, which includes the price for on-demand TPU-v3s (8 USD/hour)\footnote{\url{https://cloud.google.com/tpu/pricing}} plus auxiliary costs for virtual machines and data storage. 

We considered the possibility of evaluating checkpoints of partially pre-trained models, for fine-grained variation of the pre-training budget, however after some investigation we chose to only report results on fully pre-trained models, using established training protocols (learning rate, number of parameter updates, model size, sequence length, etc.) to ensure fair comparison.

In addition to pre-training from scratch, we also experiment with Domain-Adaptive Pre-training, using the codebase\footnote{\url{https://github.com/allenai/tpu_pretrain}} released by AI2 to train \procroberta{}. Similar to \citet{Gururangan2020DontSP}, we fine-tune \roberta{} on our collected  \textsc{Procedure} corpus for 12.5k steps with the averaged speed of 27.27 seconds per step, which leads to a TPU time of 95 hours.\footnote{This is comparable to the number reported by the authors of \citet{Gururangan2020DontSP} on \href{https://github.com/allenai/dont-stop-pretraining/issues/16}{GitHub}.} Thus, the total cost of \procroberta{} is around 800 USD after including the auxiliary expenses. 

Finally, we estimate the cost of training for 
\textbf{\scibert{}} \cite{Beltagy2019SciBERT}, which was also trained on an 8-core TPU v3 using a two-stage training process similar to \procbert{}. The overall training of \scibert{} took 7 days (5 days for the first stage and 2 days for the second stage) with an estimated cost of 1,340 USD.


\paragraph{Carbon Footprint.}
\label{sec:carbon_footprint}

Apart from the financial cost, we also estimate the carbon footprint of each in-domain pre-trained language model for its environmental impact. 
We measure the energy consumption in kilowatt-hours (KWh) as in \citet{patterson2021carbon}:
\[ Energy = H \times N \times P \times PUE / 1000, \]
\noindent where $H$ is the number of training hours, $N$ is the number of processors used, $P$ is the average power per processor,\footnote{Unlike \citet{Strubell2019EnergyAP} which measured GPU, CPU and DRAM's power separately, \citet{patterson2021carbon} measured the power of a processor together with other components including fans, network interface, host CPU, etc.} and $PUE$ (Power Usage Effectiveness) indicates the energy usage efficiency of a data center. 
In our case, the average power per TPU v3 processor is 283 watts, and we use a $PUE$ coefficient of 1.10, which is the average trailing twelve-month $PUE$ reported for all Google data centers in Q1 2021.\footnote{\url{https://www.google.com/about/datacenters/efficiency/}} 
Once we know the energy consumption, we can estimate the CO\textsubscript{2} emissions (CO\textsubscript{2}e) as follows:
\[ CO_{2}e = (Energy \times CO_{2}e/KWh) / 1000\]
\noindent where $CO_{2}e/KWh$ measures the amount of $CO_{2}$ emission when consuming 1 KWh energy, which is 474g/KWh for our pre-training.\footnote{Our models were pre-trained in the data center of Google in Netherlands: \url{https://cloud.google.com/sustainability/region-carbon}.}
For example, \procbert{} is pre-trained on a single 8-core TPU v3 for 74 hours, resulting in CO\textsubscript{2} emission of $(74\times8\times283\times1.10/1000)\times474/1000=87.4$ kg.
The estimated CO\textsubscript{2} emissions for three in-domain language models are shown in Table \ref{tab:carbon}.

\begin{table}[t!]
\small
\begin{center}
\setlength{\tabcolsep}{3pt}
\scalebox{0.85}{
\begin{tabular}{lccc}

\toprule

\textbf{Corpus}                         & \textbf{\#Tokens} & \textbf{Text Size} & \textbf{Pre-trained Model}  \\ \midrule 
Wiki + Books                   & 3.3B      & 16GB      & BERT \\ 
Web crawl                      & -         & 160GB     & RoBERTa  \\
PMC + CS                       & 3.2B      & -         & SciBERT \\
BioMed                         & 7.6B      & 47GB      & BioMed-RoBERTa  \\ \midrule
\textsc{Procedure}            & 1.05B     & 6.5GB     & Proc-RoBERTa  \\ 
\hspace{1mm} - PubMed          & 0.32B     & 2.0GB     & --  \\ 
\hspace{1mm} - Chem. patent    & 0.61B     & 3.9GB     & -- \\ 
\hspace{1mm} - Cook. recipe          & 0.11B     & 0.6GB     & --  \\ \midrule
\textsc{Procedure+}            & 12B       & 77GB      & ProcBERT  \\ 
\hspace{1mm} - \textsc{Procedure} ($\times$ 6) & 6.3B      & 39GB      & -- \\ 
\hspace{1mm} - Full articles   & 5.7B & 38GB & -- \\ 
\bottomrule 
\end{tabular}
}
\end{center}
\caption{\label{tab:corpus_size} Statistics of our newly created \textsc{Procedure} and \textsc{Procedure+} corpora, which are used for pre-training Proc-RoBERTa and ProcBERT, respectively.}
\end{table}

\begin{table}[t!]
\small
\begin{center}
\scalebox{1}{
\begin{tabular}{ll}
 
\toprule
\textbf{Consumption} & \textbf{CO\textsubscript{2}e (kg)} \\ 
\midrule
Air travel, one person,  SF$\leftrightarrow$NY & 1200\tablefootnote{Source: \href{https://support.google.com/travel/answer/9671620?hl=en}{Google Flights} \citep{patterson2021carbon}.} \\ 
\midrule
SciBERT \cite{Beltagy2019SciBERT} & 198.3\\
Proc-RoBERTa & 112.1 \\
ProcBERT & { }{ }87.4\\




\bottomrule

\end{tabular}
}
\end{center}
\caption{\label{tab:carbon} Carbon footprint of three in-domain pre-trained language models. CO\textsubscript{2}e is the number of metric tons of CO\textsubscript{2} emissions with the same global warming potential as one metric ton of another greenhouse gas.}
\end{table}

\section{Measuring Utility $U(X_a, X_p)$ under Varying Budget Constraints}
\label{sec:measure_utility}

Given the estimated unit cost of annotation $C_a$ (\S\ref{sec:estimate_c_a}) and pre-training $C_p$ (\S\ref{sec:estimate_c_p}), we now empirically evaluate the utility
$U(X_a, X_p)$, of various budgets and pre-training strategies to find an optimal policy for domain adaptation that fits within the budget constraint $X_a C_a + X_p C_P \leq B$. 


\subsection{NLP Tasks and Models}
\label{sec:utility_task}
We experiment with two NLP tasks, Named Entity Recognition (NER) and Relation Extraction (RE).
For NER, we follow \citet{devlin-etal-2019-bert} to feed the contextualized embedding of each token into a linear classification layer. 
For RE, we follow \citet{Zhong2020AFE}, inserting four special tokens specifying positions and types of each entity-pair mention, which are included as input to a pre-trained sentence encoder. 
Gold entity mentions are used in our relation extraction experiments, to reduce variance due to entity recognition errors.

\subsection{Budget-constrained Experimental Setup}
\label{sec:utility_setup}

As we have three procedural text datasets (\S \ref{sec:estimate_c_a}) annotated with entities and relations, we can experiment with six source $\Rightarrow$ target adaptation settings. For each domain pair, we compare five different pre-trained language models when adapted to the procedural text domain under varying budgets.

Based on the estimations of the annotation costs $C_a$ (\S\ref{sec:estimate_c_a}) and pre-training costs $C_p$ (\S\ref{sec:estimate_c_p}), we conduct various budget-constrained domain adaptation experiments. For example, if we have \$1,500 and the \pubmed{} corpus, to build an NER model that works best for the \chemsyn{} domain (\textsc{\pubmed{}}$\Rightarrow$\textsc{ChemSyn}), we can spend all \$1,500 to annotate 2,500 in-domain sentences to fine-tune off-the-shelf BERT. Or alternatively, we could first spend \$800 to pre-train Proc-RoBERTa, then fine-tune it on 1155 sentences annotated in the \textsc{ChemSyn} domain using the remaining of \$700. Under both budgeting strategies, an additional experiment is performed to choose one of two domain adaption methods that maximizes performance: 1) a model that simply uses the annotated data in the target domain for fine-tuning; or 2) a model which is fine-tuned using a variant of \easyadapt{} \citep{Daum2007FrustratinglyED} to leverage annotated data in both the source and target domains (see below for details). We select the approach that has better development set performance and report its test set result in Table \ref{tab:ner_results} and \ref{tab:re_results} (see Appendix \ref{sec:hyper_para} for more details about hyper-parameters).

\begin{table*}[t!]
\small
\begin{center}
\resizebox{0.98\textwidth}{!}{\begin{tabular}{ccC{0.8cm}C{0.8cm}C{0.8cm}C{0.8cm}C{0.8cm}C{0.8cm}C{0.8cm}C{0.8cm}C{0.8cm}C{0.8cm}C{0.8cm}C{0.8cm}}

\toprule 

& &  \multicolumn{2}{c}{\textbf{\bertbase{}}} & \multicolumn{2}{c}{\textbf{\bertlarge{}}} & \multicolumn{2}{c}{\textbf{\procbert{}}} & \multicolumn{2}{c}{\textbf{\procroberta{}}} & \multicolumn{2}{c}{\textbf{\scibert{}}} \\ 
\multirow{2}{*}{\textbf{Source $\Rightarrow$ Target Domain}} & \multirow{2}{*}{\textbf{Budget}} & \multicolumn{2}{c}{\$0} & \multicolumn{2}{c}{\$0} & \multicolumn{2}{c}{\$620} & \multicolumn{2}{c}{\$800} & \multicolumn{2}{c}{\$1340} \\ \cmidrule(l){3-4}\cmidrule(l){5-6}\cmidrule(l){7-8}\cmidrule(l){9-10} \cmidrule(l){11-12}

&  & F\textsubscript{1} & (\#sent) & F\textsubscript{1} & (\#sent)  & F\textsubscript{1} & (\#sent)  & F\textsubscript{1} & (\#sent)  & F\textsubscript{1} & (\#sent) \\
\midrule
\multirow{3}{*}{\textsc{PubMedM\&M}$\Rightarrow$\textsc{ChemSyn}}  
 & \$700  & 92.99\textsubscript{0.2}\textsuperscript{\textdagger} & (1166) & \textbf{93.32}\textsubscript{0.3}\textsuperscript{\textdagger} & (1166) & 91.07\textsubscript{0.4}\textsuperscript{\textdagger} & (133) & \multicolumn{2}{c}{N/A} & \multicolumn{2}{c}{N/A} \\ 
 & \$1500 & 93.58\textsubscript{0.3}\textsuperscript{\textdagger} & (2500) & 94.14\textsubscript{0.2}\textsuperscript{\textdagger} & (2500) & \textbf{94.73}\textsubscript{0.1}\textsuperscript{\textdagger} & (1466) & 93.83\textsubscript{0.1}\textsuperscript{\textdagger} & (1166) & 92.46\textsubscript{0.2}\textsuperscript{\textdagger} & (266) \\ 
 & \$2300 & 94.47\textsubscript{0.2} & (3833) & 94.81\textsubscript{0.1}\textsuperscript{\textdagger} & (3833) & \textbf{95.17}\textsubscript{0.1}\textsuperscript{\textdagger} & (2800) & 94.71\textsubscript{0.2}\textsuperscript{\textdagger} & (2500) & 94.39\textsubscript{0.2} & (1600) \\  \midrule
 
\multirow{3}{*}{\textsc{WLP}$\Rightarrow$\textsc{ChemSyn}} 
 & \$700  & 93.31\textsubscript{0.3}\textsuperscript{\textdagger} & (1166) & \textbf{93.59}\textsubscript{0.3}\textsuperscript{\textdagger} & (1166) & 90.83\textsubscript{0.3}\textsuperscript{\textdagger} & (133) & \multicolumn{2}{c}{N/A} & \multicolumn{2}{c}{N/A} \\ 
 & \$1500 & 94.31\textsubscript{0.3}\textsuperscript{\textdagger} & (2500) & 94.44\textsubscript{0.2}\textsuperscript{\textdagger} & (2500) & \textbf{94.88}\textsubscript{0.1}\textsuperscript{\textdagger} & (1466) & 94.20\textsubscript{0.2}\textsuperscript{\textdagger} & (1166) & 92.39\textsubscript{0.2}\textsuperscript{\textdagger} & (266) \\ 
 & \$2300 & 94.47\textsubscript{0.2} & (3833) & 95.09\textsubscript{0.3}\textsuperscript{\textdagger} & (3833) & \textbf{95.52}\textsubscript{0.1}\textsuperscript{\textdagger} & (2800) & 94.63\textsubscript{0.3} & (2500) & 94.39\textsubscript{0.2} & (1600)\\  \midrule
 
\multirow{2}{*}{\textsc{WLP}$\Rightarrow$\textsc{PubMedM\&M}} 
 & \$700  & 74.41\textsubscript{0.8}\textsuperscript{\textdagger} & (686) & \textbf{75.41}\textsubscript{0.5}\textsuperscript{\textdagger} & (686) & 68.30\textsubscript{0.7}\textsuperscript{\textdagger} & (78) & \multicolumn{2}{c}{N/A} & \multicolumn{2}{c}{N/A} \\ 
 & \$1500 & \multicolumn{2}{c}{N/A} & \multicolumn{2}{c}{N/A} & \textbf{76.76}\textsubscript{0.7}\textsuperscript{\textdagger} & (862) & 76.02\textsubscript{0.5} & (686) & 72.66\textsubscript{0.4}\textsuperscript{\textdagger} & (156)\\ \midrule
 
\multirow{2}{*}{\textsc{ChemSyn}$\Rightarrow$\textsc{PubMedM\&M}} 
 & \$700  & 75.10\textsubscript{0.5}\textsuperscript{\textdagger} & (686) & \textbf{75.71}\textsubscript{0.4}\textsuperscript{\textdagger} & (686) & 72.85\textsubscript{0.6}\textsuperscript{\textdagger} & (78) & \multicolumn{2}{c}{N/A} & \multicolumn{2}{c}{N/A} \\ 
 & \$1500 & \multicolumn{2}{c}{N/A} & \multicolumn{2}{c}{N/A} & \textbf{77.62}\textsubscript{0.5}\textsuperscript{\textdagger} & (862) & 76.28\textsubscript{0.5}\textsuperscript{\textdagger} & (686) & 73.93\textsubscript{0.4}\textsuperscript{\textdagger} & (156) \\ \midrule
 
\multirow{3}{*}{\textsc{ChemSyn}$\Rightarrow$\textsc{WLP}} 
 & \$700  & 72.23\textsubscript{0.4}\textsuperscript{\textdagger} & (1590) & \textbf{73.21}\textsubscript{0.2}\textsuperscript{\textdagger} & (1590) & 72.42\textsubscript{0.4}\textsuperscript{\textdagger} & (181) & \multicolumn{2}{c}{N/A} & \multicolumn{2}{c}{N/A} \\ 
 & \$1500 & 73.30\textsubscript{0.4}\textsuperscript{\textdagger} & (3409) & 73.46\textsubscript{0.4} & (3409) & \textbf{75.15}\textsubscript{0.4}\textsuperscript{\textdagger} & (2000) & 73.90\textsubscript{0.5}\textsuperscript{\textdagger} & (1590) & 72.48\textsubscript{0.4}\textsuperscript{\textdagger} & (363) \\ 
 & \$2300 & 73.18\textsubscript{0.4} & (5227) & 74.12\textsubscript{0.2}\textsuperscript{\textdagger} & (5227) & \textbf{75.88}\textsubscript{0.3}\textsuperscript{\textdagger} & (3818) & 74.67\textsubscript{0.4}\textsuperscript{\textdagger} & (3409) & 74.68\textsubscript{0.5}\textsuperscript{\textdagger} & (2181) \\  \midrule
 
\multirow{3}{*}{\textsc{PubMedM\&M}$\Rightarrow$\textsc{WLP}} 
 & \$700  & 72.66\textsubscript{0.3}\textsuperscript{\textdagger} & (1590) & \textbf{73.18}\textsubscript{0.7}\textsuperscript{\textdagger} & (1590) & 72.91\textsubscript{0.3}\textsuperscript{\textdagger} & (181) & \multicolumn{2}{c}{N/A} & \multicolumn{2}{c}{N/A} \\ 
 & \$1500 & 73.62\textsubscript{0.4}\textsuperscript{\textdagger} & (3409) & 73.46\textsubscript{0.4} & (3409) & \textbf{75.25}\textsubscript{0.1}\textsuperscript{\textdagger} & (2000) & 73.98\textsubscript{0.2}\textsuperscript{\textdagger} & (1590) & 72.58\textsubscript{0.2}\textsuperscript{\textdagger} & (363) \\ 
 & \$2300 & 73.73\textsubscript{0.3}\textsuperscript{\textdagger} & (5227) & 74.26\textsubscript{0.2}\textsuperscript{\textdagger} & (5227) & \textbf{75.78}\textsubscript{0.3}\textsuperscript{\textdagger} & (3818) & 74.68\textsubscript{0.2}\textsuperscript{\textdagger} & (3409) & 74.80\textsubscript{0.2}\textsuperscript{\textdagger} & (2181)\\ \bottomrule 
\end{tabular}}
\end{center}
\caption{\label{tab:ner_results} Experiment results for Named Entity Recognition (NER). With higher budgets (\$1500 and \$2300), our in-domain pre-training of \procbert{} achieves the best results in combination with data annotation. For a smaller budget (\$700), investing all funds in annotation and fine-tuning the standard \bertlarge{} (considered as cost-free) will yield the best outcome. \textbf{\#sent} is the number of sentences from the target domain, annotated under the given budget, used for training. \textsuperscript{\textdagger} indicates results using EasyAdapt (\S \ref{sec:utility_easy}), where source domain data helps. }
\end{table*}

\subsection{\easyadapt{}}
\label{sec:utility_easy}
In most of our experiments, 
we have access to a relatively large amount of labeled data from a source domain, and varying amounts of data from the target domain. Instead of simply concatenating the source and target datasets for fine-tuning, we propose a simple, yet novel variation of EasyAdapt \citep{Daum2007FrustratinglyED} for pre-trained Transformers. More specifically, we create three copies of the model's contextualized representations: one represents the source domain, one represents the target, and the third is domain-independent. These contextualized vectors are then concatenated and fed into a linear layer that is 3 times as large as the base model's.  When encoding data from a specific domain (e.g. \chemsyn{}), the other domain's representations are zeroed out (1/3 of the new representations will always be 0.0). This enables the domain-specific block of the linear layer to encode information specific to that domain, while the domain-independent parameters can learn to represent information that transfers across domains. This is similar to prior work using EasyAdapt \citep{kim2016frustratingly} for LSTMs.

\subsection{Experimental Results and Analysis}
\label{sec:utility_result}

We present the test set NER and RE results with five annotation and pre-training combination strategies under six domain adaptation settings in Table \ref{tab:ner_results} and \ref{tab:re_results}, respectively. We report averages across five random seeds with standard deviations as subscripts. If pre-training costs go over the total budget, or available data goes under the annotation budget, we indicate the result as \quotes{NA}.  We now discuss a set of key questions regarding pre-training-based domain adaptation under a constrained budget, which our experiments can shed some light on.

\paragraph{Should we prioritize pre-training or data annotation for domain adaptation?} For all six domain adaptation settings in Table \ref{tab:ner_results}, spending the entire budget on annotation and using the off-the-shelf language model \bertlarge{} works the best for NER when the budget is 700 USD, showing the effectiveness of data annotation in low-resource scenarios. 
As the budget increases, performance gains from labelling additional data diminish, and pre-training in-domain language models takes the lead.
\procbert{}, which is pre-trained from scratch on the {\sc Procedure+} corpus costing only 620 USD, performs best at budgets of 1500 and 2300 USD. 
This demonstrates that combining domain-specific pre-training with data annotation is the best strategy in high-resource settings.
Similarly for RE, as shown in Table \ref{tab:re_results}, using all funds for data annotation and working with off-the-shelf models achieves better performance at lower budgets, while domain-specific pre-training starts to excel as the budget increases past a certain point.

\begin{table*}[t!]
\small
\begin{center}
\resizebox{0.98\textwidth}{!}{\begin{tabular}{ccC{0.8cm}C{0.8cm}C{0.8cm}C{0.8cm}C{0.8cm}C{0.8cm}C{0.8cm}C{0.8cm}C{0.8cm}C{0.8cm}C{0.8cm}C{0.8cm}}

\toprule 

& &  \multicolumn{2}{c}{\textbf{\bertbase{}}} & \multicolumn{2}{c}{\textbf{\bertlarge{}}} & \multicolumn{2}{c}{\textbf{\procbert{}}} & \multicolumn{2}{c}{\textbf{\procroberta{}}} & \multicolumn{2}{c}{\textbf{\scibert{}}} \\ 
\multirow{2}{*}{\textbf{Source $\Rightarrow$ Target Domain}} & \multirow{2}{*}{\textbf{Budget}} & \multicolumn{2}{c}{\$0} & \multicolumn{2}{c}{\$0} & \multicolumn{2}{c}{\$620} & \multicolumn{2}{c}{\$800} & \multicolumn{2}{c}{\$1340} \\ \cmidrule(l){3-4}\cmidrule(l){5-6}\cmidrule(l){7-8}\cmidrule(l){9-10} \cmidrule(l){11-12}

&  & F\textsubscript{1} & (\#sent) & F\textsubscript{1} & (\#sent)  & F\textsubscript{1} & (\#sent)  & F\textsubscript{1} & (\#sent)  & F\textsubscript{1} & (\#sent) \\
\midrule
\multirow{3}{*}{\textsc{PubMedM\&M}$\Rightarrow$\textsc{ChemSyn}}  
 & \$700  & 90.12\textsubscript{0.4}\textsuperscript{\textdagger} & (1166) & \textbf{90.71}\textsubscript{0.8}\textsuperscript{\textdagger} & (1166) & 88.21\textsubscript{0.6}\textsuperscript{\textdagger} & (133) & \multicolumn{2}{c}{N/A} & \multicolumn{2}{c}{N/A}\\ 
 & \$1500 & 91.30\textsubscript{0.2}\textsuperscript{\textdagger} & (2500) & 91.84\textsubscript{0.5}\textsuperscript{\textdagger} & (2500) & \textbf{92.06}\textsubscript{0.4}\textsuperscript{\textdagger} & (1466) & 91.25\textsubscript{0.7}\textsuperscript{\textdagger} & (1166) & 89.58\textsubscript{0.5}\textsuperscript{\textdagger} & (266) \\ 
 & \$2300 & 91.81\textsubscript{0.2}\textsuperscript{\textdagger} & (3833) & \textbf{92.90}\textsubscript{0.4}\textsuperscript{\textdagger} & (3833) & 92.57\textsubscript{0.2}\textsuperscript{\textdagger} & (2800) & 92.07\textsubscript{0.3}\textsuperscript{\textdagger} & (2500) & 91.55\textsubscript{0.2}\textsuperscript{\textdagger} & (1600) \\  \midrule
 
\multirow{3}{*}{\textsc{WLP}$\Rightarrow$\textsc{ChemSyn}} 
 & \$700  & \textbf{90.20}\textsubscript{0.7}\textsuperscript{\textdagger} & (1166) & 90.15\textsubscript{0.9} & (1166) & 88.01\textsubscript{0.6}\textsuperscript{\textdagger} & (133) & \multicolumn{2}{c}{N/A} & \multicolumn{2}{c}{N/A}\\ 
 & \$1500 & 91.34\textsubscript{0.3}\textsuperscript{\textdagger} & (2500) & 91.61\textsubscript{0.5}\textsuperscript{\textdagger} & (2500) & \textbf{92.27}\textsubscript{0.2}\textsuperscript{\textdagger} & (1466) & 91.77\textsubscript{0.2}\textsuperscript{\textdagger} & (1166) & 89.16\textsubscript{0.7}\textsuperscript{\textdagger} & (266) \\ 
 & \$2300 & 92.08\textsubscript{0.4}\textsuperscript{\textdagger} & (3833) & 92.73\textsubscript{0.4}\textsuperscript{\textdagger} & (3833) & \textbf{92.85}\textsubscript{0.2}\textsuperscript{\textdagger} & (2800) & 92.44\textsubscript{0.6}\textsuperscript{\textdagger} & (2500) & 91.42\textsubscript{0.4}\textsuperscript{\textdagger} & (1600) \\  \midrule
 
\multirow{2}{*}{\textsc{WLP}$\Rightarrow$\textsc{PubMedM\&M}} 
 & \$700 & 77.74\textsubscript{0.7}\textsuperscript{\textdagger} & (686) & \textbf{79.33}\textsubscript{1.3}\textsuperscript{\textdagger} & (686) & 74.63\textsubscript{0.4}\textsuperscript{\textdagger} & (78) & \multicolumn{2}{c}{N/A} & \multicolumn{2}{c}{N/A} \\ 
 & \$1500 & \multicolumn{2}{c}{N/A} & \multicolumn{2}{c}{N/A} & \textbf{80.10}\textsubscript{0.6}\textsuperscript{\textdagger} & (862) & 79.33\textsubscript{1.3}\textsuperscript{\textdagger} & (686) & 75.70\textsubscript{1.1}\textsuperscript{\textdagger} & (156) \\ 
 \midrule
 
\multirow{2}{*}{\textsc{ChemSyn}$\Rightarrow$\textsc{PubMedM\&M}} 
 & \$700 & 77.02\textsubscript{0.6}\textsuperscript{\textdagger} & (686) & \textbf{77.31}\textsubscript{0.4}\textsuperscript{\textdagger} & (686) & 73.92\textsubscript{0.8}\textsuperscript{\textdagger} & (78) & \multicolumn{2}{c}{N/A} & \multicolumn{2}{c}{N/A} \\ 
 & \$1500 & \multicolumn{2}{c}{N/A} & \multicolumn{2}{c}{N/A} & \textbf{79.50}\textsubscript{0.7}\textsuperscript{\textdagger} & (862) & 77.12\textsubscript{1.2}\textsuperscript{\textdagger} & (686) & 75.43\textsubscript{0.6}\textsuperscript{\textdagger} & (156) \\ 
 \midrule
 
\multirow{3}{*}{\textsc{ChemSyn}$\Rightarrow$\textsc{WLP}} 
  & \$700 & 78.81\textsubscript{0.6}\textsuperscript{\textdagger} & (1590) & \textbf{79.80}\textsubscript{0.6}\textsuperscript{\textdagger} & (1590) & 78.97\textsubscript{0.7}\textsuperscript{\textdagger} & (181) & \multicolumn{2}{c}{N/A} & \multicolumn{2}{c}{N/A} \\ 
 & \$1500 & 79.91\textsubscript{0.4}\textsuperscript{\textdagger} & (3409) & 80.15\textsubscript{0.8}\textsuperscript{\textdagger} & (3409) & \textbf{80.87}\textsubscript{0.7}\textsuperscript{\textdagger} & (2000) & 79.69\textsubscript{0.4}\textsuperscript{\textdagger} & (1590) & 79.44\textsubscript{0.8}\textsuperscript{\textdagger} & (363) \\ 
 & \$2300 & 80.54\textsubscript{0.6}\textsuperscript{\textdagger} & (5227) & 80.97\textsubscript{0.2}\textsuperscript{\textdagger} & (5227) & 81.15\textsubscript{0.6}\textsuperscript{\textdagger} & (3818) & \textbf{81.33}\textsubscript{0.5}\textsuperscript{\textdagger} & (3409) & 80.16\textsubscript{0.7}\textsuperscript{\textdagger} & (2181) \\  \midrule
 
\multirow{3}{*}{\textsc{PubMedM\&M}$\Rightarrow$\textsc{WLP}} 
  & \$700 & 78.34\textsubscript{0.8}\textsuperscript{\textdagger} & (1590) & \textbf{78.44}\textsubscript{1.2}\textsuperscript{\textdagger} & (1590) & 77.98\textsubscript{0.8}\textsuperscript{\textdagger} & (181) & \multicolumn{2}{c}{N/A} & \multicolumn{2}{c}{N/A} \\ 
 & \$1500 & 79.40\textsubscript{0.6}\textsuperscript{\textdagger} & (3409) & 79.78\textsubscript{0.6}\textsuperscript{\textdagger} & (3409) & \textbf{80.45}\textsubscript{0.2}\textsuperscript{\textdagger} & (2000) & 78.79\textsubscript{1.1}\textsuperscript{\textdagger} & (1590) & 78.76\textsubscript{0.5}\textsuperscript{\textdagger} & (363) \\ 
 & \$2300 & 79.93\textsubscript{0.5}\textsuperscript{\textdagger} & (5227) & 80.04\textsubscript{0.8}\textsuperscript{\textdagger} & (5227) & \textbf{80.85}\textsubscript{0.6}\textsuperscript{\textdagger} & (3818) & 79.49\textsubscript{0.7}\textsuperscript{\textdagger} & (3409) & 80.33\textsubscript{0.3}\textsuperscript{\textdagger} & (2181) \\  
 \bottomrule 
\end{tabular}}
\end{center}
\caption{\label{tab:re_results} Experiment results for Relation Extraction (RE). Similar to the observations in Table \ref{tab:ner_results}, regardless of a small or large budget, prioritizing data annotation in the target domain is the most beneficial. \textsuperscript{\textdagger} indicates results using EasyAdapt (\S \ref{sec:utility_easy}), where source domain data helps. }
\end{table*}

\begin{figure*}[t!]
    \centering
    \begin{subfigure}[t]{0.5\textwidth}
        \centering
        \includegraphics[width=8cm]{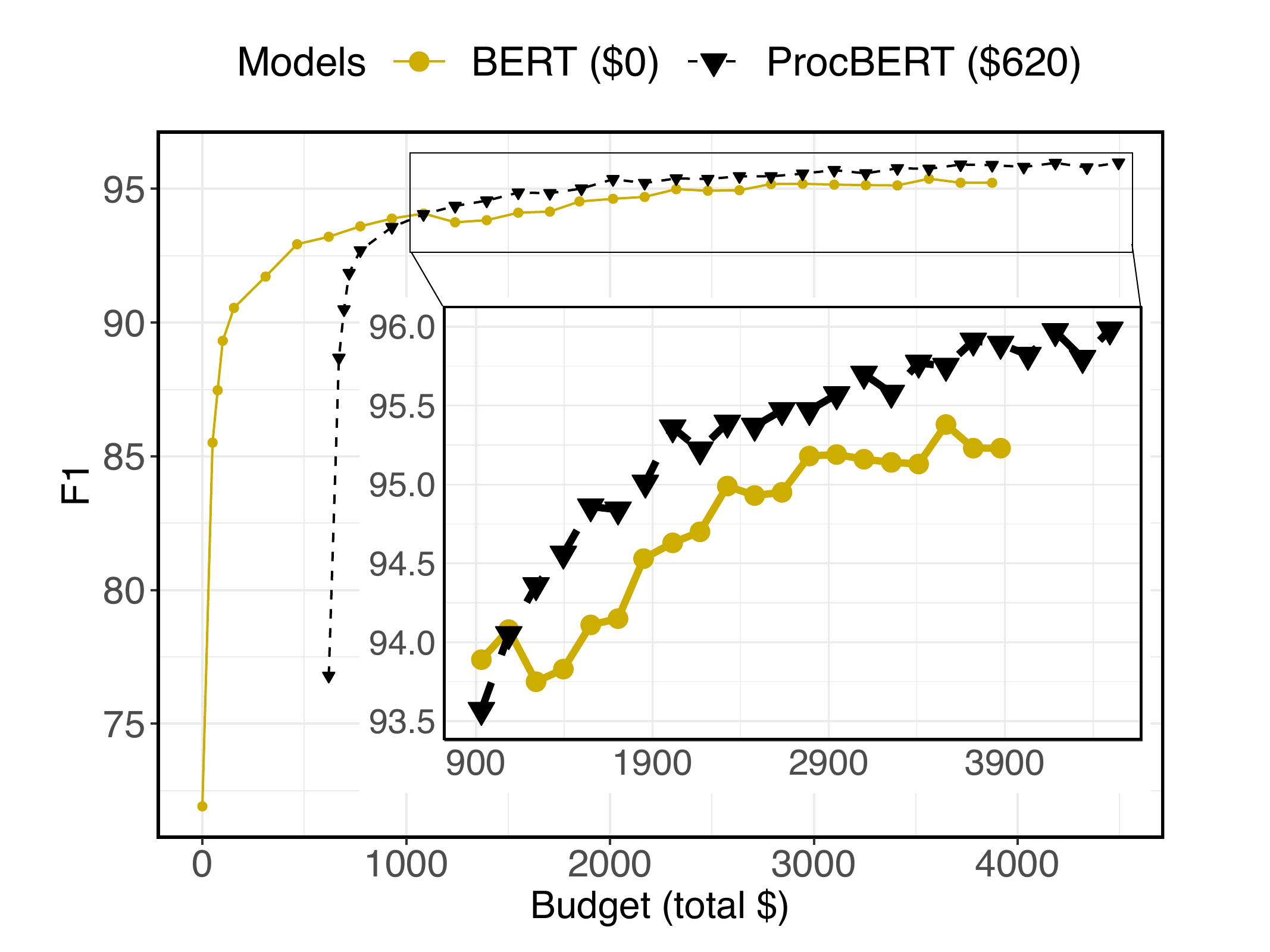}
        \caption{\pubmed{}$\Rightarrow$\chemsyn{}}
    \end{subfigure}%
    ~ 
    \begin{subfigure}[t]{0.5\textwidth}
        \centering
        \includegraphics[width=8cm]{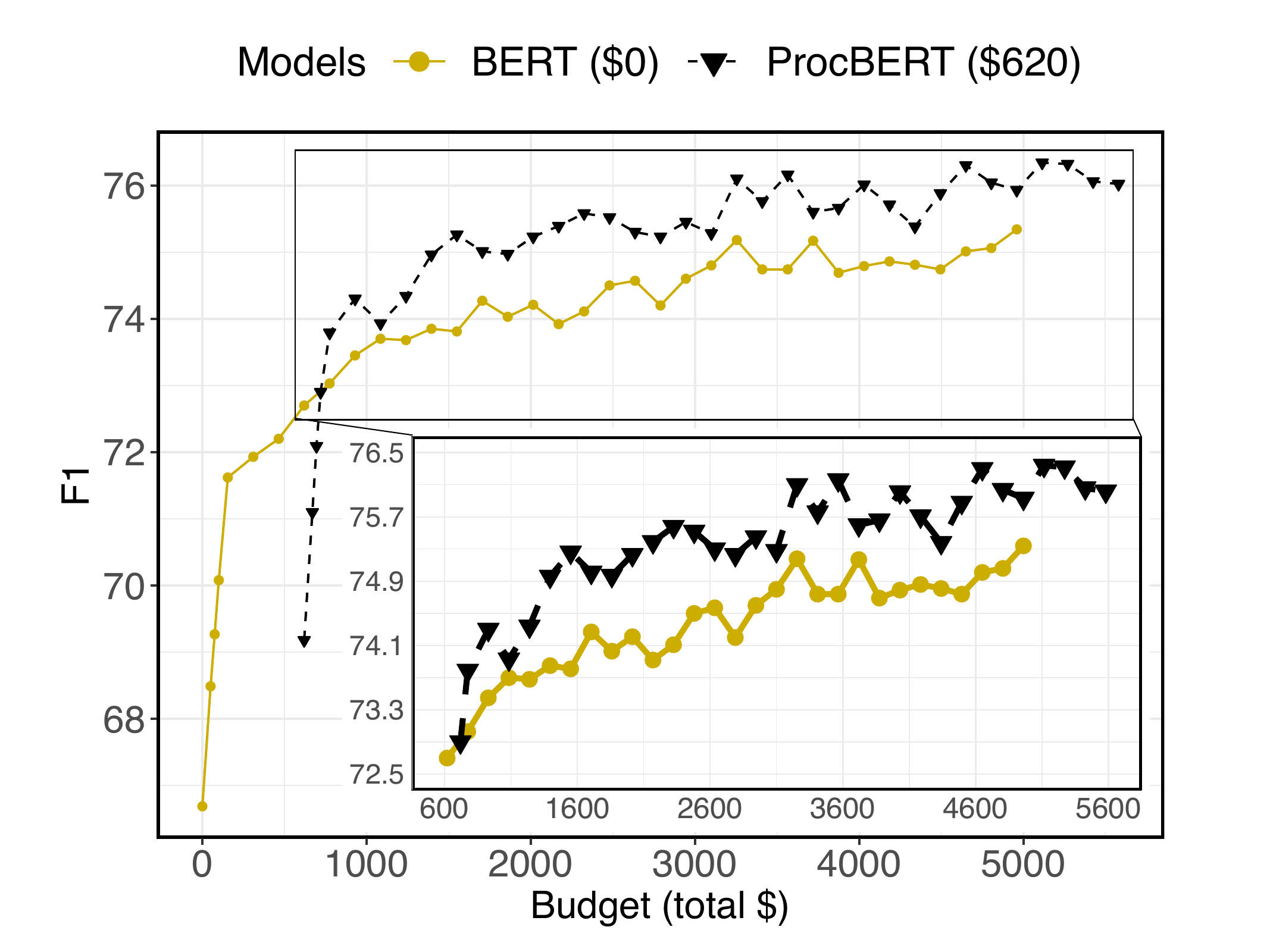}
        \caption{\chemsyn{}$\Rightarrow$\wlp{}}
    \end{subfigure}
    \caption{\label{fig:budget} Comparison of two domain adaptation strategies: 1) \emojibert allocate all available funds to data annotation; 2)  \emojiprocbert pre-train \procbert{} on in-domain data, then use the remaining budget for annotation.  For small budgets, the former yields the best performance on  NER, but as the budget increases, the later becomes the best choice.}
\end{figure*}

\paragraph{What is the starting budget to consider pre-training an in-domain language model?}
To answer this question, 
we plot test set NER performance for two strategies, \bertlarge{} (investing all funds on annotation) and \procbert{} (combining annotation with pre-training), against varying budgets in Figure \ref{fig:budget}. Specifically, the budget of each strategy starts with the pre-training cost of its associated language model, and is increased by 155 USD increments until the total budget reaches the total cost of available data\footnote{We also add a few points at the start of each curve to make them smoother. For \bertlarge{}, we add 50 USD, 75 USD and 100 USD. For \procbert{}, we add 670 USD, 695 USD and 720 UDS.}.
We observe a similar trend for both \pubmed{}$\Rightarrow$\chemsyn{} and \chemsyn{}$\Rightarrow$\wlp{}: annotation alone works better in lower budgets while in-domain pre-training (\procbert{}) excels at higher budgets.
However, the intersection of the curves for these two strategies occurs at different points, which are around 1085 USD and 775 USD.

\begin{figure*}[t!]
    \centering
    \begin{subfigure}[t]{0.5\textwidth}
        \centering
        \includegraphics[width=8cm]{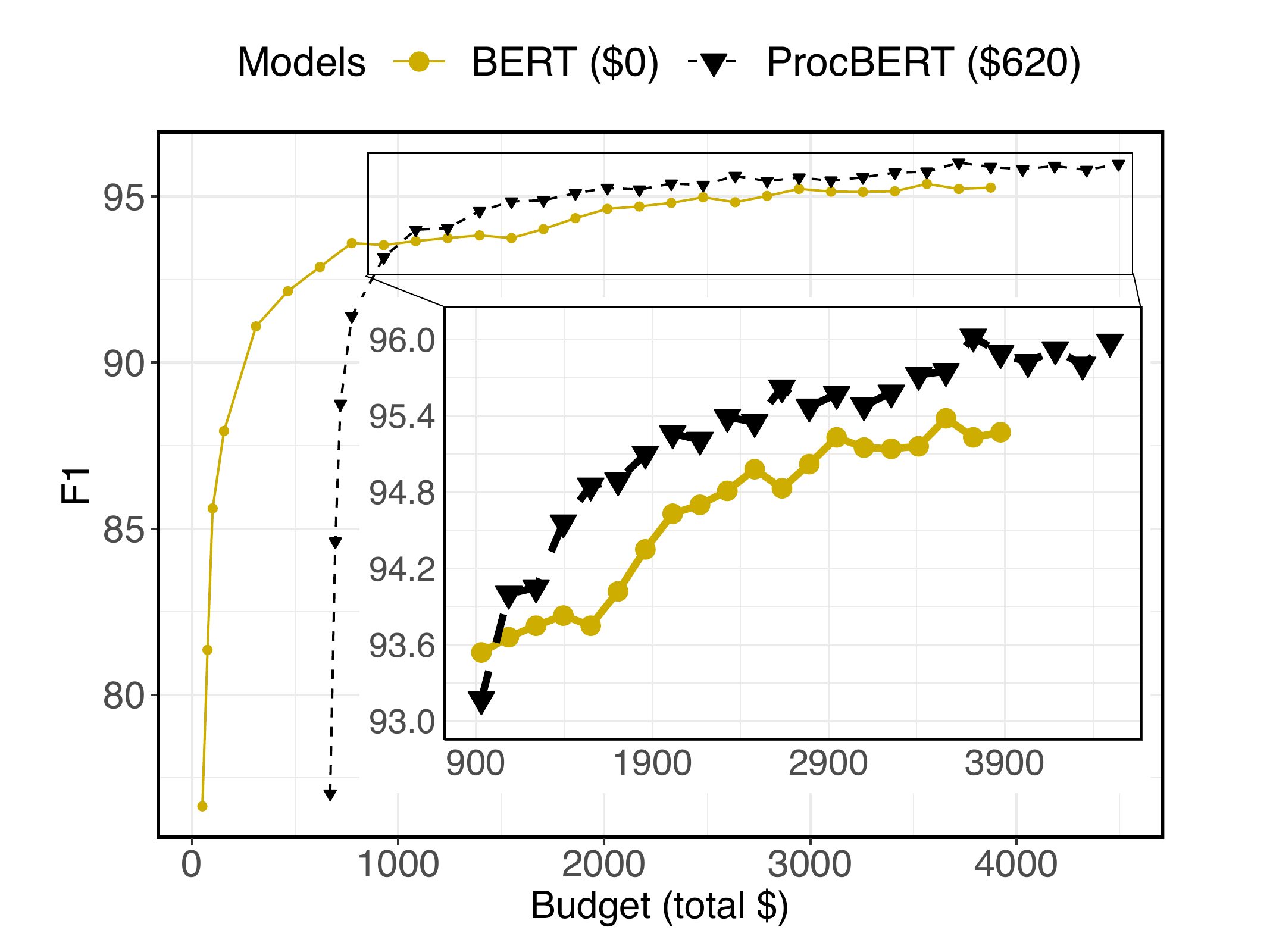}
        \caption{\chemsyn{}}
    \end{subfigure}%
    ~ 
    \begin{subfigure}[t]{0.5\textwidth}
        \centering
        \includegraphics[width=8cm]{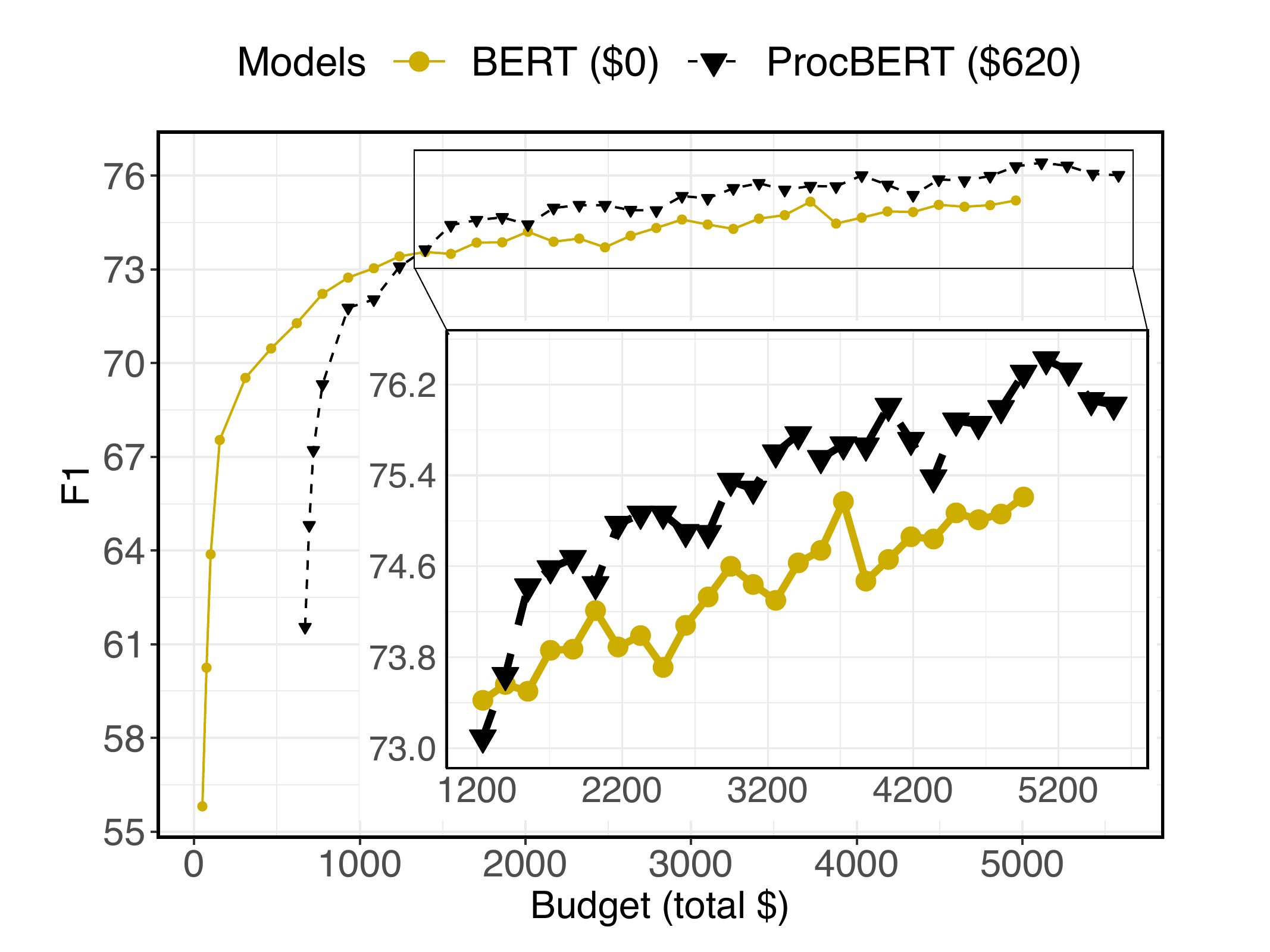}
        \caption{\wlp{}}
    \end{subfigure}
    \caption{\label{fig:budget_tgt} Comparison of spending the entire budget on data annotation (\emojibert) and pre-training followed by in-domain annotation (\emojiprocbert), where models are trained on \textbf{target domain labeled data only}. The crossover point for \wlp{} moves from 775 USD (adapted from \chemsyn{}) to around 1395 USD (\wlp{} only) demonstrating that a large source domain dataset can reduce the need for target domain annotation.}
\end{figure*}



We hypothesize there are two reasons for this difference: 1) each target domain may require different amounts of labeled data to generalize well; 2) the quantity of labeled data from the source domain may also impact the need for data annotation in the target domain. 
To testify our hypotheses, we evaluate the utility of annotation vs. pre-training where no source-domain data is available in  Figure \ref{fig:budget_tgt}.  This is almost identical to the setting of Figure \ref{fig:budget} except models are trained on \textbf{target domain labeled data only}. Here, we observe the intersection for \chemsyn{} is still around 1085 USD while the crossover point for \wlp{} moves from the original 775 USD (in Figure \ref{fig:budget}) to around 1395 USD. Our hypothesis is that \wlp{} is a broader domain compared to \chemsyn{} (\wlp{} covers a more diverse range of protocols that include cell cultures, DNA sequencing, etc.), so it requires more annotated data to perform well under the setting of Figure \ref{fig:budget_tgt}. However, when adapted from a large source domain dataset like \chemsyn{}, the need for annotated \wlp{} corpus is reduced so that \procbert{} can outperform \bertlarge{} at a lower budget.


Note that our estimated annotation cost, $C_a$, (per sentence) includes the annotation of both entity mentions and relations, so our analysis amortizes the cost of pre-training across both tasks.  In a scenario where more tasks need to be adapted for the target domain, this could be accounted for simply by dividing the cost of pre-training among tasks, which would shift the black curves in Figure \ref{fig:budget} and Figure \ref{fig:budget_tgt} to the left, making pre-training an economical choice at lower budgets.\footnote{For more discussion on our assumptions, see \S \ref{sec:assumptions}.}

\begin{table}[t!]
\small

\begin{center}
\setlength{\tabcolsep}{4pt}
\scalebox{0.9}{
\begin{tabular}{crcc}
\toprule 
\multirow{2}{*}{\textbf{Target}} &  \multirow{2}{*}{\textbf{Entities}} & \textbf{\bertlarge{}} & \textbf{\procbert{}} \\
 & & \$0 & \$620  \\
\midrule
 & seen by both & \textbf{97.63} & 97.45 \\
\chemsyn{} &  unseen by both & 84.96 & \textbf{85.65} \\
(budget \$1085) &  seen by BERT & \textbf{94.07} & 93.44 \\
\cmidrule{2-4}
&  All & 93.78 & 93.82 \\
\midrule
 & seen by both & \textbf{85.42} & 84.78 \\
\wlp{} &  unseen by both & 55.75 & \textbf{57.19} \\
(budget \$1395) &  seen by BERT & \textbf{76.50} & 76.27 \\
\cmidrule{2-4}
&  All & 73.51 & 73.51 \\
\bottomrule
\end{tabular}
}
\end{center}
\caption{\label{tab:anno_pretrain_compare} Test set F\textsubscript{1} on NER for entities seen and unseen in the training data for  \bertlarge{} and \procbert{}, when the two achieve very similar overall performance under the same budget constraints in Figure \ref{fig:budget_tgt}. \procbert{} performs better on the unseen entities.}
\end{table}

\paragraph{When using the same budget and achieving similar F\textsubscript{1}, how do pre-training and annotation differ?} In the previous experiments, we show that in-domain pre-training is an effective domain adaptation method especially in high-budget settings. \procbert{} can work very well when trained with less labeled data. 
A plausible explanation is that in-domain pre-training improves generalization to new entities in the target domain, whereas additional annotation improves the performance on entities that are observed in the training corpus.
To evaluate this hypothesis, we compare model predictions of the two strategies at the crossover points in Figure \ref{fig:budget_tgt}\footnote{We choose Figure \ref{fig:budget_tgt} for this analysis instead of Figure \ref{fig:budget} because we want to isolate the impact of source domain labeled data.}, and consider each entity in the test set as "Seen" or "Unseen" based on whether it was observed in the training set. Then, we calculate the F\textsubscript{1} score for each category as shown in Table \ref{tab:anno_pretrain_compare}.  Although the models that are compared achieve nearly identical overall performance, the decomposition of performance on seen and unseen entities in Table \ref{tab:anno_pretrain_compare} clearly suggests that in-domain pre-training leads to better generalization on unseen entities, whereas allocating more budget to annotation boosts performance on entities that were seen during training.  This may help provide an explanation for the main finding of this paper: in-domain pre-training results in better generalization on unseen mentions, leading to better marginal utility, but only after enough in-domain annotations are observed to fully cover the head of the distribution.

\section{Positive Externalities of Pretraining}
\label{sec:other_procedural}
\begin{table*}[th!]
\small

\begin{center}
\setlength{\tabcolsep}{3pt}
\resizebox{\textwidth}{!}{\begin{tabular}{lccccccccccc}
\toprule 
\multirow{2}{*}{\textbf{Model}}   & \multicolumn{2}{c}{\bf\textsc{X-WLP}} & \multicolumn{2}{c}{\bf\textsc{CheMU}} & \multicolumn{1}{c}{\bf\textsc{Recipe}} & \multicolumn{2}{c}{\bf\textsc{WLP}} & \multicolumn{2}{c}{\bf\textsc{PubMedM\&M}} & \multicolumn{2}{c}{\bf\textsc{ChemSyn}}  \\ 
               & \multicolumn{1}{c}{Core} & \multicolumn{1}{c}{Non-Core} & \multicolumn{1}{c}{NER} & \multicolumn{1}{c}{EE} & \multicolumn{1}{c}{ET} &  \multicolumn{1}{c}{NER} & \multicolumn{1}{c}{RE} & \multicolumn{1}{c}{NER} & \multicolumn{1}{c}{RE} & \multicolumn{1}{c}{NER}  &  \multicolumn{1}{c}{RE}   \\ \midrule

\bertbase{} & 74.79\textsubscript{0.6} & 77.57\textsubscript{0.6} & 95.14\textsubscript{0.1} & 91.93\textsubscript{0.6} & 80.62\textsubscript{0.7} & 73.87\textsubscript{0.4} & 80.25\textsubscript{0.4}    &  74.80\textsubscript{0.7} & 
78.73\textsubscript{0.9} & 95.09\textsubscript{0.2} & 92.63\textsubscript{0.2}  \\ 

\bertlarge{} & \underline{75.53\textsubscript{1.7}} & 76.77\textsubscript{0.5} & 95.10\textsubscript{0.2} & 92.10\textsubscript{0.9} & 81.53\textsubscript{0.5} & 74.97\textsubscript{0.3} & 81.39\textsubscript{0.5} & 77.06\textsubscript{0.3} & 78.44\textsubscript{1.3} & 95.26\textsubscript{0.1} & 92.87\textsubscript{0.5}                \\
                 
\robbase{} & 75.04\textsubscript{0.8} & 77.24\textsubscript{0.6} & 95.05\textsubscript{0.1} & \textbf{92.54}\textsubscript{1.2} & 83.41\textsubscript{0.1} & 74.97\textsubscript{0.5} & 80.94\textsubscript{0.5}    & 76.21\textsubscript{0.3} & 
78.95\textsubscript{0.7}    & 95.30\textsubscript{0.2} & 93.39\textsubscript{0.3}  \\ 

\roblarge{} & 73.77\textsubscript{1.6} & 74.37\textsubscript{0.2} & 95.16\textsubscript{0.2} & 92.10\textsubscript{1.1} & \textbf{84.54}\textsubscript{0.8} & \textbf{76.37}\textsubscript{0.5} & 79.76\textsubscript{0.3} & \textbf{78.70}\textsubscript{0.6} & 75.66\textsubscript{0.3} & 95.66\textsubscript{0.2} & 92.87\textsubscript{0.2}          \\
                 
\scibert{}          & 75.48\textsubscript{0.7} & \underline{78.51\textsubscript{0.6}} &  \underline{95.63\textsubscript{0.0}} & 91.80\textsubscript{0.5} & 81.75\textsubscript{0.4} &  75.89\textsubscript{0.4} & 81.29\textsubscript{0.5}    & \underline{77.81\textsubscript{0.2}} & 
79.54\textsubscript{0.9}    & 95.82\textsubscript{0.2} & 93.27\textsubscript{0.2}  \\ 

\biomedrob{}   & 74.89\textsubscript{0.6} & 76.39\textsubscript{0.8} & 95.32\textsubscript{0.2} & \underline{92.42\textsubscript{0.6}} & 82.92\textsubscript{0.3} & 75.55\textsubscript{0.2} & \textbf{81.56}\textsubscript{ 0.4}   & 77.14\textsubscript{0.1} & 
78.44\textsubscript{2.0} & 95.38\textsubscript{0.2} & 93.16\textsubscript{0.3} \\ \midrule

\procroberta{} & 74.76\textsubscript{0.7} & 76.12\textsubscript{0.9} & 95.49\textsubscript{0.1} & 91.55\textsubscript{0.6} & \underline{84.19\textsubscript{0.3}} & 75.76\textsubscript{0.4} & 80.79\textsubscript{0.9}    &   76.91\textsubscript{0.7}  & 
79.16\textsubscript{0.9}    & 95.67\textsubscript{0.1} & 93.31\textsubscript{0.2}      \\

\procbert{} & \textbf{76.73}\textsubscript{0.9} & \textbf{78.57}\textsubscript{0.8} & \textbf{96.19\textsubscript{0.1}} & 92.32\textsubscript{0.2} & 84.10\textsubscript{0.3} & \underline{76.04\textsubscript{0.2}} & \underline{81.44\textsubscript{0.4}}    & 77.31\textsubscript{0.5} & 
\textbf{80.19}\textsubscript{0.6} & \textbf{95.97}\textsubscript{0.2} & \textbf{93.57}\textsubscript{0.2}   \\

\midrule

\textsc{SOTA} &  76.5 &  78.1 & 95.70 & 95.36 & 81.96 & 77.99 & 80.46            & -- & -- & -- & -- \\ 
 \bottomrule
\end{tabular}}
\end{center}
\vspace{-.2cm}
\caption{\label{tab:pretrain_results} Test set F\textsubscript{1} on six procedural text datasets. The best task performance is boldfaced, and the second-best performance is underlined.
For the SOTA model of each dataset, we refer readers to the corresponding paper for further details: \citet{tamari-etal-2021-process-level} for \xwlp{}, \citet{Wang2020MelaxtechAR} for \chemu{}, \citet{gupta-durrett-2019-effective} for \recipe{}, \citet{knafou-etal-2020-bitem} for NER on \wlp{}, and \citet{Sohrab2020mgsohrabAW} for RE on \wlp{}.
}
\end{table*}

So far, we have discussed domain adaptation as a consumer choice problem
where annotation and pre-training costs are balanced to maximize performance in a target domain.  However, pre-training on large quantities of natural language instructions can improve performance on additional tasks in the procedural text domain, as demonstrated in the following subsections.

\subsection{Ancillary Procedural NLP Tasks}

In addition to the procedural text datasets discussed in \S \ref{sec:measure_utility}, we experiment with three ancillary procedural text corpora, to explore how in-domain pretraining can benefit other tasks.

The \textbf{\chemu{} corpus} \citep{Nguyen2020ChEMUNE} contains NER and event annotations for 1500 chemical reaction snippets collected from 170 English patents. 
Its NER task focuses on identifying chemical compounds, and its event extraction (EE) task aims at detecting chemical reaction events including trigger detection and argument role labeling.

The \textbf{\xwlp{} corpus} \citep{tamari-etal-2021-process-level} provides the Process Event Graphs (PEG) of 279 wet-lab biochemistry protocols. The PEG is a document-level graph-based representation specifying the involved experimental objects.

The \textbf{\recipe{} corpus} \citep{Kiddon2016GloballyCT}  includes annotation of entity states for 866 cooking recipes. It supports Entity Tracking (ET) task which predicts whether or not a specific ingredient is involved in each step of the recipe.

\subsection{Experiments on Ancillary Tasks}
For \chemu{}, gold arguments are provided, so we only need to identify the event trigger and predict the role of the gold arguments.  An event prediction is correct if the event trigger, associated arguments, and their roles match with the gold event mention. We tackle this task using a pipeline model similar to \citet{Zhong2020AFE}.
For \xwlp{}, we focus on the operation argument role labeling task, where gold entities are provided as input.  
Following \citet{tamari-etal-2021-process-level}, we decompose the results into "Core" and "Non-Core" roles.
For the \recipe{} task, we follow the data splits and fine-tuning architecture of \citet{gupta-durrett-2019-effective}. The state of an ingredient in each cooking step is correct if it matches with the gold labels, as either present or absent.
\paragraph{Results.}
Test set results of eight pre-trained language models on six procedural text datasets are presented in Table \ref{tab:pretrain_results}.\footnote{For \chemu{}, we report the development set results because its test set is not publicly available.} \procbert{}, performs best in most tasks and even achieves the state-off-the-art performance on operational argument role labeling ("Core" and "Non-Core") of \xwlp{}, showing the effectiveness of in-domain pre-training. 


\section{Conclusion}
\label{sec:conclu}
In this paper, we address a number of questions related to the costs of adapting an NLP model to a new domain \citep{blitzer2006domain,Han2019UnsupervisedDA}, an important and well-studied problem in NLP.  We frame domain adaptation under a constrained budget as a problem of consumer choice.  Experiments are conducted using several pre-trained models in three procedural text domains to determine when it is economical to pre-train in-domain transformers \cite{Gururangan2020DontSP}, and when it is better to spend available resources on annotation.  Our results suggest that when a small number of NLP models need to be adapted to a new domain, pre-training, by itself, is not an economical solution.

\section*{Acknowledgments}
We are grateful to the anonymous reviewers for helpful feedback on an earlier draft of this paper.  We also thank Vardaan Pahuja for assistance with extracting experimental paragraphs from PubMed, and John Niekrasz for sharing the output of Daniel Lowe's reaction extraction tool on European patents.
This material is based upon work supported by the Defense Advanced Research Projects Agency (DARPA) under Contract No. HR001119C0108, in addition to the NSF (IIS-1845670) and IARPA via the BETTER program (2019-19051600004).
The views, opinions, and/or findings expressed are those of the author(s) and should not be interpreted as representing the official views or policies of the Department of Defense, IARPA or the U.S. Government.  This work is approved for Public Release, Distribution Unlimited.


\bibliography{anthology,custom}
\bibliographystyle{acl_natbib}

\appendix



\newpage

\section{Data Annotation}
\label{sec:data_anno}

We annotate two datasets \textsc{PubMedM\&M} and \textsc{ChemSyn} in the domain of scientific articles and chemical patents mainly following the annotation scheme of the Wet Lab Protocols \cite[WLP;][]{Tabassum2020WNUT2020T1}. On top of 20 entity types and 16 relation types in WLP, we supplement four entity types (\textsc{Company}, \textsc{Software}, \textsc{Data-Collection} and \textsc{Info-Type}) and one relation type (\textsc{Belong-To}) due to two key features of our corpus: 1) scientific articles usually specify the provenance of reagents for better reproducibility; 
2) it covers a broader range of procedures such as computer simulation and data analysis.

We recruit four undergraduate students to annotate the datasets using the BRAT annotation tool.\footnote{\url{https://github.com/nlplab/brat}} 
We double-annotate all files in \textsc{PubMedM\&M} and half of the files in \textsc{ChemSyn}. For those double-annotated files, the coordinator will discuss the annotation with each annotator making sure their annotation follows the guideline and dissolve the disagreement. As for the inter-annotator agreement (IAA) score, we treat the annotation from one of the two annotator as the gold label, and the other annotation as the predicted label, and then use the F1 scores of Entity(Action) and Relation evaluations as the final inter-annotator agreement scores, which are shown in Table \ref{tab:iaa_results}. 
We can see that \textsc{ChemSyn} has higher IAA scores, and there are two potential reasons: 1) we annotate \textsc{PubMedM\&M} first, so the annotators might be more experienced when they annotate \textsc{ChemSyn}; 2) \textsc{PubMedM\&M} contains more diverse content like wet lab experiments or computer simulation procedures while \textsc{ChemSyn} is mainly about chemical synthesis.



\begin{table}[h!]
\small
\begin{center}
\begin{tabular}{ccc}
\toprule
Dataset & Entities/Actions & Relations  \\ \midrule
\textsc{PubMedM\&M} & 70.54 & 51.94 \\

\textsc{ChemSyn} & 79.87 & 87.20 \\ 

\bottomrule
\end{tabular}
\end{center}
\caption{\label{tab:iaa_results} Inter-Annotator Agreement (F\textsubscript{1} scores on Entity/Action Identification and Relation Extraction).}
\end{table}

\section{Procedural Corpus Collection}
\label{sec:app_corpus}
\paragraph{PubMed Articles.}
The first source of our procedural corpus is PubMed articles
because they contain a large number of freely accessible experimental procedures.
Specifically, we extract procedural paragraphs 
from the {\em Materials and Methods} section of articles within the Open Access Subset of PubMed.  XML files containing full text of articles are downloaded from NCBI\footnote{\url{https://www.ncbi.nlm.nih.gov/}} and then processed to obtain all the paragraphs within the {\em Materials and Methods} section.

To improve the quality of our collected corpus, we develop a procedural paragraph extractor by fine-tuning SciBERT \citep{Beltagy2019SciBERT} on the SciSeg dataset \citep{Dasigi2017ExperimentSI}, which includes discourse labels ({\tt \{Goal, Fact, Result, Hypothesis, Method, Problem, Implication\}}) for PubMed articles.
This extractor achieves an average F1 score of 72.65\% in a  five-fold cross validation, and we run it on all acquired 
paragraphs.
We consider a paragraph as a valid procedure if at least 40\% of clauses are labeled as {\tt Method}.  This threshold is obtained by manual inspection of the randomly sampled subset of the data.

In total, the PubMed Open Access Subset contains 2,542,736 articles, of which about 680k contain a {\em Materials and Methods} section.  After running our trained procedural paragraph extractor, we retain a set of 1,785,923 procedural paragraphs.  Based on a manual inspection of the extracted paragraphs, we estimate that 92\% consist of instructions for carrying out experimental procedures.

\paragraph{Chemical Patents.}
The second source of our corpus is the patent data because chemical patents usually include detailed procedures of chemical synthesis. 
We download U.S. patent data (1976-2016) from USPTO\footnote{\url{https://www.uspto.gov/learning-and-resources/bulk-data-products}} and European data (1978-2020) from EPO\footnote{\url{https://www.epo.org/searching-for-patents/data/bulk-data-sets.html}} as XML files. 
Then we apply the reaction extractor developed by \citet{Lowe2012ExtractionOC}, a trained Naive Bayes classifier, to the {\em Description} section of our collected patents. Note that the U.S. patent data has two subsets, "Grant" (1976-2016) and "Application" (2001-2016). The "Application" subset covers the "Grant" subset from the same year, so for those overlapping years (2001-2016), we only use the U.S. patents from the "Application" subset. 
As a result, we get 2,435,999 paragraphs from 174,554 U.S. patents and 4,039,6065 paragraphs from 129,035 European patents. 
Lastly, we use the language identification tool \textbf{langid}\footnote{\url{https://github.com/saffsd/langid.py}} to build an English-only corpus, which includes 6,107,481 paragraphs.


\section{Pre-training Details}

We pre-train \procbert{} using the TensorFlow codebase of BERT \citep{devlin-etal-2019-bert}.\footnote{\url{https://github.com/google-research/bert}} We use the Adam optimizer \citep{Kingma2015AdamAM} with $\beta_1$ = 0.9, $\beta_2$ = 0.999 and $L2$ weight decay of 0.01. Following \citet{devlin-etal-2019-bert}, we deploy the two-step regime. In the first step, we pre-train the model with sequence length 128 and batch size 512 for 1 million steps. The learning rate is warmed up over the first 100k steps to a peak value of 1e-4, then linearly decayed. In the second step, we train 100k more steps of sequence length 512 and batch size 256 to learn the positional embeddings with peak learning rate 2e-5. We use the original sub-word mask as the masking strategy, and we mask 15\% of tokens in the sequence for both training steps.

For \procroberta{}, we use the codebase from AI2,\footnote{\url{https://github.com/allenai/tpu_pretrain}} which enables language model pre-training on TPUs with PyTorch. Similar to \citet{Gururangan2020DontSP}, we train \roberta{} on our collected procedural text corpus for 12.5k steps with a learning rate of 3e-5 and an effective batch size 2048, which is achieved by accumulating the gradient of 128 steps with a basic batch size of 16. The input sequence length is 512 throughout the whole process, and 15\% of words are masked for prediction.

\section{Hyper-parameters for Downstream Tasks}
\label{sec:hyper_para}

We use the same five random seeds as \citet{peters-etal-2019-knowledge} for all our experiments in \S\ref{sec:measure_utility} and \S\ref{sec:other_procedural}.\footnote{\url{https://github.com/allenai/kb/blob/master/bin/run_hyperparameter_seeds.sh}} 
We select the best hyperparameter values based on the average development set performances over five random seeds by grid search.
For models with \bertbase{} or \robbase{} architecture, the search range includes learning rate (1e-5, 2e-5), batch size (16, 48, 64, 128), max sequence length (128, 256, 512) and epoch number (5, 20, 60), and the used hyperparameter values on budget-constrained domain adaptation experiments (denoted as "\textsc{Budget}") (\S\ref{sec:measure_utility}) and ancillary tasks (\S\ref{sec:other_procedural}) are shown in Table \ref{tab:hyper_base}. 
For \bertlarge{} and \roblarge{}, the search range is different in learning rate (5e-6, 1e-5), batch size (4, 8, 12, 24, 64) and epoch number (3, 5, 10, 20), and the used values are shown in Table \ref{tab:hyper_large}.


\begin{table*}[t!]
\small
\begin{center}
\setlength{\tabcolsep}{3pt}
\resizebox{\textwidth}{!}{\begin{tabular}{lC{1.2cm} C{1.2cm}C{1.2cm}C{1.2cm}C{1.2cm}C{1.2cm}C{1.2cm}C{1.2cm}C{1.2cm}C{1.2cm}C{1.2cm}C{1.2cm}}
\toprule 
\multirow{2}{*}{\textbf{Hyperparam.}} & \multicolumn{2}{c}{\bf\textsc{Budget}} & \bf\textsc{X-WLP} & \multicolumn{2}{c}{\bf\textsc{CheMU}} & \multicolumn{1}{c}{\bf\textsc{Recipe}} & \multicolumn{2}{c}{\bf\textsc{WLP}} & \multicolumn{2}{c}{\bf\textsc{PubMedM\&M}} & \multicolumn{2}{c}{\bf\textsc{ChemSyn}}  \\ 
            & \multicolumn{1}{c}{NER} & \multicolumn{1}{c}{RE}  & ARL & \multicolumn{1}{c}{NER} & \multicolumn{1}{c}{EE} & \multicolumn{1}{c}{ET} &  \multicolumn{1}{c}{NER} & \multicolumn{1}{c}{RE} & \multicolumn{1}{c}{NER} & \multicolumn{1}{c}{RE} & \multicolumn{1}{c}{NER}  &  \multicolumn{1}{c}{RE}   \\
\midrule

Learning Rate  & 1e-5 & 1e-5 & 1e-5 & 1e-5 & 1e-5 & 2e-5 & 1e-5 & 1e-5 & 1e-5 & 1e-5 & 1e-5 & 1e-5 \\
Batch Size & 16 & 48 & 16 & 16 & 16 & 16 & 16 & 128 & 16 & 64 & 16 & 48 \\
Max Seq. Length & 512 & 256 & 512 & 512 & 512 & 512 & 256 & 128 & 512 & 256 & 512 & 256\\
Max Epoch & 20 & 5 & 5 & 20 & 5 & 20 & 20 & 5 & 60 & 5 & 20 & 5\\

 \bottomrule
\end{tabular}}
\end{center}
\caption{\label{tab:hyper_base} Hyperparameters for models with \bertbase{} or \robbase{} architecture on budget-constrained domain adaptation experiments (denoted as "\textsc{Budget}") (\S\ref{sec:measure_utility}) and ancillary tasks (\S\ref{sec:other_procedural}).
}
\end{table*}

\begin{table*}[t!]
\small
\begin{center}
\setlength{\tabcolsep}{3pt}
\resizebox{\textwidth}{!}{\begin{tabular}{lC{1.2cm}C{1.2cm}C{1.2cm}C{1.2cm}C{1.2cm}C{1.2cm}C{1.2cm}C{1.2cm}C{1.2cm}C{1.2cm}C{1.2cm}C{1.2cm}}
\toprule 
\multirow{2}{*}{\textbf{Hyperparam.}} & \multicolumn{2}{c}{\bf\textsc{Budget}} & \bf\textsc{X-WLP} & \multicolumn{2}{c}{\bf\textsc{CheMU}} & \multicolumn{1}{c}{\bf\textsc{Recipe}} & \multicolumn{2}{c}{\bf\textsc{WLP}} & \multicolumn{2}{c}{\bf\textsc{PubMedM\&M}} & \multicolumn{2}{c}{\bf\textsc{ChemSyn}}  \\ 
            & \multicolumn{1}{c}{NER} & \multicolumn{1}{c}{RE}  & ARL & \multicolumn{1}{c}{NER} & \multicolumn{1}{c}{EE} & \multicolumn{1}{c}{ET} &  \multicolumn{1}{c}{NER} & \multicolumn{1}{c}{RE} & \multicolumn{1}{c}{NER} & \multicolumn{1}{c}{RE} & \multicolumn{1}{c}{NER}  &  \multicolumn{1}{c}{RE}   \\
\midrule

Learning Rate  & 1e-5 & 5e-6 & 5e-6 & 1e-5 & 5e-6 & 5e-6 & 1e-5 & 5e-6 & 1e-5 & 5e-6 & 1e-5 & 5e-6 \\
Batch Size & 4 & 12 & 4 & 4 & 4 & 4 & 4 & 64 & 4 & 24 & 4 & 12\\
Max Seq. Length & 512 & 256 & 512 & 512 & 512 & 512 & 256 & 128 & 512 & 256 & 512 & 256\\
Max Epoch & 10 & 3 & 3 & 10 & 3 & 5 & 10 & 3 & 20 & 5 & 10 & 3\\

 \bottomrule
\end{tabular}}
\end{center}
\caption{\label{tab:hyper_large} Hyperparameters for \bertlarge{} and \roblarge{} on budget-constrained domain adaptation experiments (\S\ref{sec:measure_utility}) and ancillary tasks (\S\ref{sec:other_procedural}).
}
\end{table*}

\end{document}